\documentclass[11pt]{article}
\usepackage{coling2020}
\usepackage{times}
\usepackage{url}
\usepackage{latexsym}
\usepackage{comment}
\usepackage[colorinlistoftodos]{todonotes}
\newcommand{\eg}{{\textit{e.g.}}}

\usepackage{xcolor}
\usepackage{multirow}

\colingfinalcopy % Uncomment this line for the final submission

\title{MaintNet: A Collaborative Open-Source Library for Predictive Maintenance Language Resources}

\author{Farhad Akhbardeh, Travis Desell, Marcos Zampieri  \\
  Rochester Institute of Technology, United States \\
  \{\tt fa3019, tjdvse, mazgla\}@rit.edu \\}

\date{}

\begin{document}
\maketitle

\begin{abstract}
\label{sec:abstract}
Maintenance record logbooks are an emerging text type in NLP. They typically consist of free text documents with many domain specific technical terms, abbreviations, as well as non-standard spelling and grammar, which poses difficulties to NLP pipelines trained on standard corpora. Analyzing and annotating such documents is of particular importance in the development of predictive maintenance systems, which aim to provide operational efficiencies, prevent accidents and save lives. In order to facilitate and encourage research in this area, we have developed MaintNet, a collaborative open-source library of technical and domain-specific language datasets. MaintNet provides novel logbook data from the aviation, automotive, and facilities domains along with tools to aid in their (pre-)processing and clustering. Furthermore, it provides a way to encourage discussion on and sharing of new datasets and tools for logbook data analysis.
\end{abstract}

\section{Introduction}
\label{sec:introduction}
With the rapid development of information technologies, engineering systems are generating ever increasing amounts of data 
that are used by industries to improve products. Maintenance records are one of these types of data. They typically consist of event logbooks which are collected in many domains such as aviation, transportation, and healthcare~\cite{Tanguy2016NaturalLP,Altuncu2018FromTT}. The analysis of maintenance records is particularly important in the development of predictive maintenance systems, which can be used to improve efficiencies as well as prevent accidents and reduce maintenance costs~\cite{Jarry2018AircraftAA}. 

Maintenance record datasets generally contain free text fields describing issues (or problems) and actions, as can be seen in the instances presented in Table \ref{tab:AviationDatasets}.

\begin{table*}[ht!]
\centering
\scalebox{0.86}{
\begin{tabular}{cp{7.5cm}cp{6.8cm}}
\hline
\bf ID	& \bf	Issue/Problem	& \bf	Date	& \bf	Action	\\ \hline
111552	& R/H FWD UPPER BAFL SEAL NEEDS TO BE RESECURED & 7/2/2012 & INSTALLED POP RIVET TO RESECURE R/H FWD BAF SEEAL.\\\hline
111563	& CAP SCREWE MISSING, L/H ENG \#4 BAFLE& 7/3/2012	& INSTALLED NEW SCREW. CHKD ENG\\\hline
111574	& CYL \#1 BAFFLE CRACKED AT SCREW SUPPORT \& FWD BAFL BELOWE \#1& 7/2/2012& FABRICATED PATCHES OF LIKE MATERIAL \& RIVETED IAW CESSN\\\hline
111585	& \#3 FWD PUSH ROD TUBE GSK LEAKING @ EGNINE& 7/2/2012	& REMOVED \& REPLACED \#3 FWD PUSH ROD TUBE SEALS. LEAK CHE\\
\hline
\end{tabular}
}
\caption{Four examples from Maintnet's aviation dataset.}
\vspace{-2mm}
\label{tab:AviationDatasets}
\end{table*}

\noindent Standard NLP tools, however, are typically trained on standard contemporary corpora (e.g. newspaper texts) and struggle when dealing with the domain specific terminology, abbreviations, and non-standard spelling which are abundant in maintenance records. To help encourage further study in this area, we present MaintNet\footnote{Available at: \url{ https://people.rit.edu/fa3019/MaintNet/}}, a collaborative, open-source library for technical language resources with a special focus on predictive maintenance data.

The four main contributions of this paper are the following: 

\begin{enumerate}
    \item The development of MaintNet, a user-friendly web-based platform that serves as a repository hosting a variety of resources related to predictive maintenance and technical logbook data.
    \vspace{-2mm}
    \item The creation of several important language resources for technical language and predictive maintenance such as abbreviation lists, morphosyntactic information lists, and termbanks for the aviation, automotive and facilities domains. All these resources as well as raw data from these domains are made freely available to the research community via MaintNet.
    \vspace{-2mm}
    \item The development of several novel Python packages for (pre-)processing technical language including stop word removal, stemmers, lemmatizers, POS tagging, clustering methods, and more.
    \vspace{-2mm}
    \item A collaborative environment in which the community can contribute with data and resources and interact with developers and other members of the community. 
\end{enumerate}

\section{MaintNet Features}
\label{sec:COLTproposedSystem}

\subsection{Language Resources}
\label{sec:data}

To the best of our knowledge, there are no freely available tools and libraries developed to process such data which makes MaintNet unique. MaintNet currently features datasets from the aviation, automotive, and facilities domains (see Table \ref{tab:datasets}), and it will be expanded with the collaboration of the interested members of the NLP community working on similar topics.

\begin{table*}[ht!]
\centering
\scalebox{0.90}{
\begin{tabular}{llllp{7.8cm}}
\hline
\textbf{Domain} & \textbf{Dataset} & \textbf{Inst.} & \textbf{Tokens} & \textbf{Source} \\
\hline
Automotive & Car & 617 & 4,443 & Connecticut Open Data \\ \hline
Aviation &  Maintenance & 6,169 & 76,866 & University of North Dakota Aviation Program \\
  & Accident & 5,268 & 162,533 & Open Data by Socrata\\
  & Safety & 25,558 & 345,979 & Federal Aviation Administration \\ \hline
Facility Maintenance & Operations & 87,276 & 2,469,003 & Baltimore City Maryland Preventive Maintenance\\
\hline
\end{tabular}
}
\caption{The number of instances and tokens in each dataset/domain.}
\label{tab:datasets}
\end{table*}

\noindent Predictive maintenance datasets are hard to obtain due to the sensitive information they contain. Therefore, we work closely with the data providers to ensure that any confidential and sensitive information in the dataset remains anonymous. In addition to the datasets, MaintNet further provides the user with domain specific abbreviation dictionaries, morphosyntactic annotation, and term banks. The abbreviation dictionaries contains abbreviated validated by domain experts. The morphosyntactic annotation contains the part of speech (POS) tag, compound, lemma, and word stems. Finally, the domain term banks contain the collected list of terms that are used in each domain along with a sample of usage in the corpus.   

\subsection{Pre-processing and Tools}
\label{sec:preprocess}

\begin{figure*}[htbp!]
    \centering
    \includegraphics[height=50mm, width=160mm, scale=0.6]{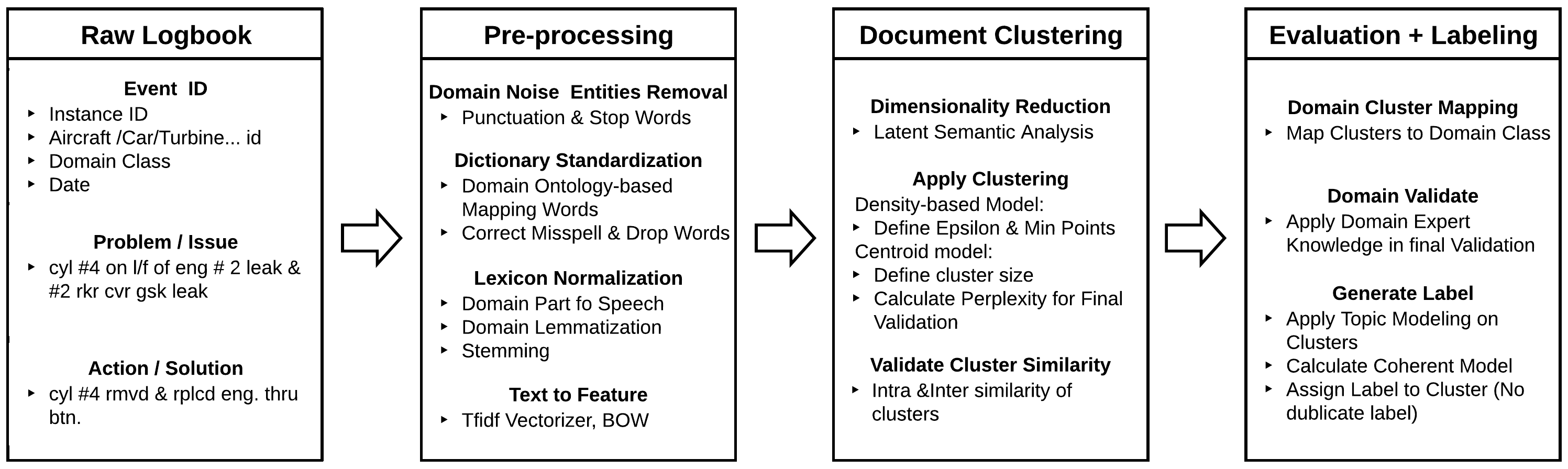}
    \vspace{-6mm}
    \caption{A pipeline of pre-processing and information extraction of maintenance dataset in MaintNet.}
    \vspace{-2mm}
    \label{fig:preprocess}
\end{figure*}

One of the bottlenecks of automatically processing logbooks for predictive maintenance is that most of these datasets are not annotated with the reason of maintenance or the category of the issue. We implemented several (pre-)processing steps to clean and extract as much information from logbooks as possible. The pipeline is shown in Figure~\ref{fig:preprocess}.  The process starts with text normalization, including lowercasing, stop word and punctuation removal, and treating special characters with NLTK's \cite{Bird2009NaturalLP} regular expression library, followed by stemming (Snowball Stemmer), lemmatization (WordNet \cite{Miller1992WORDNETAL}), and tokenization (NLTK tokenizer). With use of the collected morphosyntactic information, POS annotation is carried out with the NLTK POS tagger. Term frequency-inverse document frequency (TF-IDF) is obtained using the \emph{gensim tfidf model} \cite{Rehurek2010SoftwareFF}. Our analysis of the logbooks found that many of the misspellings and abbreviations lead to incorrect or non-existent dictionary look ups. To overcome this issue, we explored various state-of-the-art spellcheckers including Enchant\footnote{\url{https://www.abisource.com/projects/enchant/}}, Pyspellchecker\footnote{\url{https://github.com/barrust/pyspellchecker}}, Symspellpy\footnote{\url{https://github.com/wolfgarbe/SymSpell}}, and Autocorrect\footnote{\url{https://github.com/fsondej/autocorrect}}.

\vspace{-2mm}

\noindent Given the inaccuracy of existing techniques, we developed methods of correcting syntactic errors, typos, and abbreviated words using a Levenshtein distance-based algorithm \cite{Aggarwal2012ASO}. This method uses a dictionary of domain specific words and maps the various possible misspelled words into the correct format by selecting the most similar word in the dictionary. The Levenshtein algorithm was chosen over other distance metrics (\eg, Euclidian, Cosine) as it allows us to control the minimum number of string edits. The results of our method compared to other spellchecking techniques in a sub set of the aviation dataset is presented in Table \ref{tab:spellchecker_algorithm}.
\vspace{2mm}

\begin{table}[hbt!]
\begin{center}
\scalebox{0.82}{
\begin{tabular}{ p{5cm} p{3cm} p{1cm} }
 \hline
 \textbf{Total Number of Documents} & &200 \\ [0.5ex] 
 \hline
 \textbf{Total Unique Tokens} & &106  \\ 
 \hline
 \textbf{Total Non-standard} & &97  \\ 
 \hline 
    \multirow{5}{*}{\textbf{Success Rate (\%)}}&Enchant       &59\\
                                              &PySpellchecker&12\\
                                              &Autocorrect   &45\\
                                              &Levenshtein   &97\\[1ex] 
 \hline
 \end{tabular}
 }
 \end{center}
 \vspace{-2mm}
 \caption{\label{font-table} Results of the spelling correction and abbreviation expansion methods.}
 \label{tab:spellchecker_algorithm}
 \end{table}

\noindent WordNet was used to lemmatize the document, however it requires defining a POS tagger parameter which we want to lemmatize (the wordNet default is ``noun''). As the maintenance instances typically consist of verb, noun, adverb and adjective words that define a problem, action and occurrence, by using ``verb'' as the POS parameter, there is an issue of mapping important noun words such as ``left'' (\eg~left engine) to ``leave'' or ``ground'' to ``grind''. To resolve this issue, as we discussed in \ref{sec:data}, we created an exception list using developed morphosyntactic  information for the WordNet lemmatizer to ignore mapping words which could be multiple parts of speech. We convert the terms and words into a numerical representation using libraries such as \emph{tfidfvectorizer} \cite{ElSahar2017UnsupervisedOR} resulting in a large matrix of document terms (DT). 

MaintNet also features implementations of popular clustering algorithms applied to logbook data that are made freely available to the research community. The motivation behind this is that most of this data is not annotated, which requires a domain expert to group instances into categories. Clustering techniques were used to help in this process. We use truncated singular value decomposition (SVD) \cite{ElSahar2017UnsupervisedOR} known as latent semantic analysis (LSA), to perform a linear dimensionality reduction. We chose truncated SVD (LSA) over principal component analysis (PCA)\cite{ElSahar2017UnsupervisedOR} in our system, due to the fact LSA can directly be applied to our \emph{tfidf} DT matrix and it focuses on document and term relationships where PCA focuses on a term covariance matrix (eigendecomposition of the correlation). We experimented with different 4 clustering techniques: k-means \cite{Jain2010DataC5}, Density-Based Spatial Clustering of Applications with Noise (DBSCAN) \cite{Ester1996ADA}, Latent Dirichlet Analysis (LDA) \cite{Vorontsov2015BigARTMOS}, and hierarchical clustering \cite{Aggarwal2012ASO}. For comparison of the results, the silhouette and inertia \cite{Fraley1998HowMC} metrics were used to determine the number of clusters for k-means (both provided similar results), and perplexity \cite{Fraley1998HowMC} and coherence \cite{Vorontsov2015BigARTMOS} scores were used for LDA. DBSCAN and hierarchical clustering do not require a predetermined number of clusters. 

For evaluation, we used a standard measurement of cluster cohesion including high intra-cluster similarity  and low inter-cluster similarity. We chose 3 different similarity algorithms including Levenshtein, Jaro, and cosine \cite{Fraley1998HowMC} to calculate intra- and inter-cluster similarity. The cosine similarity metric is commonly used and is independent of the length of document, while Jaro is more flexible by providing a rating of matching strings. We collected human annotated instances by a domain expert to serve as our gold standard, and these are provided on MaintNet to encourage research into improving unsupervised clustering of maintenance logbooks.

\subsection{Community Participation}
\label{sec:participation}
\begin{figure*}[htbp!]
    \centering
    \includegraphics[scale=0.255]{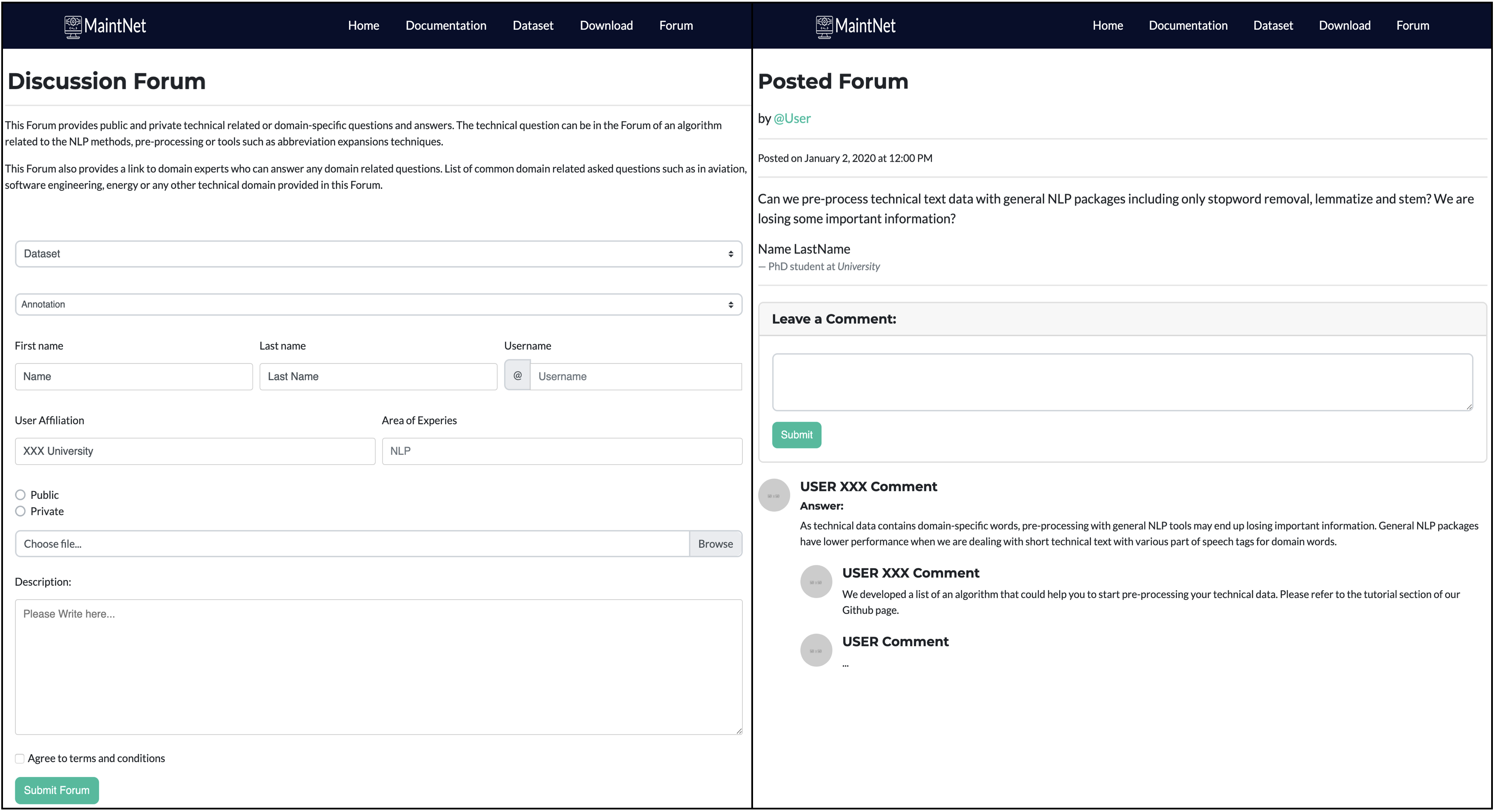}
    \vspace{-2mm}
    \caption{A screenshot of MaintNet's discussion webpages.}
    \label{fig:form}
\end{figure*}

Finally, MaintNet provides various webpages for users to communicate with each other and the project developers; as well as upload data to share with the community (see Figure~\ref{fig:form}). We hope this will help further facilitate discussion and research in this under explored area.

\section{Conclusions and Future Work}
\label{sec:conclusions}

In this paper we presented MaintNet, a collaborative open-source library for predictive maintenance language resources. MaintNet provides raw technical logbook data as well as several language resources such as abbreviation lists, morphosyntactic information lists, and termbanks from the aviation, automotive and facilities domains. Tools developed in Python are also made available for pre-processing, such as spell checking, POS tagging, and document clustering. In addition to these tools, the collaborative aspects of MaintNet should be emphasized. We welcome the community to contribute with new datasets that can be processed using the tools available at MaintNet, or share new and improved tools developed with MaintNet's open source data. 

MaintNet is also expanding as current work involves processing data from additional domains such as healthcare and power systems ({\it e.g.}, wind turbines). These datasets will be made available on MaintNet in upcoming months. We also aim to collect and release datasets and tools for languages other than English in the near future.

%\newpage
% \section*{Acknowledgements}

\bibliographystyle{coling}
\bibliography{coling2020}

\begin{thebibliography}{}

\bibitem[\protect\citename{Aggarwal and Zhai}2012]{Aggarwal2012ASO}
Charu~C. Aggarwal and ChengXiang Zhai.
\newblock 2012.
\newblock A survey of text clustering algorithms.
\newblock In {\em Mining Text Data}.

\bibitem[\protect\citename{Altuncu \bgroup et al.\egroup
  }2018]{Altuncu2018FromTT}
M.~Tarik Altuncu, Erik Mayer, Sophia~N. Yaliraki, and Mauricio Barahona.
\newblock 2018.
\newblock From text to topics in healthcare records: An unsupervised graph
  partitioning methodology.
\newblock {\em ArXiv}, abs/1807.02599.

\bibitem[\protect\citename{Bird \bgroup et al.\egroup }2009]{Bird2009NaturalLP}
Steven Bird, Ewan Klein, and Edward Loper.
\newblock 2009.
\newblock {\em Natural Language Processing with Python}.
\newblock O'Reilly.

\bibitem[\protect\citename{ElSahar \bgroup et al.\egroup
  }2017]{ElSahar2017UnsupervisedOR}
Hady ElSahar, Elena Demidova, Simon Gottschalk, Christophe Gravier, and
  Fr{\'e}d{\'e}rique Laforest.
\newblock 2017.
\newblock Unsupervised open relation extraction.
\newblock {\em ArXiv}, abs/1801.07174.

\bibitem[\protect\citename{Ester \bgroup et al.\egroup }1996]{Ester1996ADA}
Martin Ester, Hans-Peter Kriegel, J{\"o}rg Sander, and Xiaowei Xu.
\newblock 1996.
\newblock A density-based algorithm for discovering clusters in large spatial
  databases with noise.
\newblock In {\em KDD}.

\bibitem[\protect\citename{Fraley and Raftery}1998]{Fraley1998HowMC}
Chris Fraley and Adrian~E. Raftery.
\newblock 1998.
\newblock How many clusters? which clustering method? answers via model-based
  cluster analysis.
\newblock {\em Comput. J.}, 41:578--588.

\bibitem[\protect\citename{Jain}2010]{Jain2010DataC5}
Anil~Kumar Jain.
\newblock 2010.
\newblock Data clustering: 50 years beyond k-means.
\newblock {\em Pattern Recognition Letters}, 31:651--666.

\bibitem[\protect\citename{Jarry \bgroup et al.\egroup
  }2018]{Jarry2018AircraftAA}
Gabriel Jarry, Daniel Delahaye, Florence Nicol, and Eric Feron.
\newblock 2018.
\newblock Aircraft atypical approach detection using functional principal
  component analysis.
\newblock In {\em SID}.

\bibitem[\protect\citename{Miller}1992]{Miller1992WORDNETAL}
George~A. Miller.
\newblock 1992.
\newblock Wordnet: A lexical database for english.
\newblock {\em Commun. ACM}, 38:39--41.

\bibitem[\protect\citename{Rehurek and Sojka}2010]{Rehurek2010SoftwareFF}
Radim Rehurek and Petr Sojka.
\newblock 2010.
\newblock Software framework for topic modelling with large corpora.
\newblock In {\em LREC}.

\bibitem[\protect\citename{Tanguy \bgroup et al.\egroup
  }2016]{Tanguy2016NaturalLP}
Ludovic Tanguy, Nikola Tulechki, Assaf Urieli, Eric Hermann, and C{\'e}line
  Raynal.
\newblock 2016.
\newblock Natural language processing for aviation safety reports: From
  classification to interactive analysis.
\newblock {\em Computers in Industry}, 78:80--95.

\bibitem[\protect\citename{Vorontsov \bgroup et al.\egroup
  }2015]{Vorontsov2015BigARTMOS}
Konstantin Vorontsov, Oleksandr Frei, Murat Apishev, Peter Romov, and Marina
  Dudarenko.
\newblock 2015.
\newblock Bigartm: Open source library for regularized multimodal topic
  modeling of large collections.
\newblock In {\em AIST}.

\end{thebibliography}

\end{document}